\documentclass[conference,10pt,letterpaper]{IEEEtran}
\IEEEoverridecommandlockouts

\usepackage[utf8]{inputenc}

\usepackage[caption=false,font=footnotesize]{subfig}
\usepackage{booktabs}

\usepackage{url}
\usepackage{amsmath,amsfonts,amsthm}
\usepackage{listings}
\usepackage{xspace}
\usepackage{fnpct}
\usepackage{tikz}
\usepackage{tikzscale}
\usepackage{tabularx}
\newcolumntype{R}{>{\raggedleft\arraybackslash}X}

\usepackage{pgf,pgfplots,pgfplotstable,forest,tikz}
\usepgfplotslibrary{external,groupplots}
\usetikzlibrary{positioning}
\usetikzlibrary{pgfplots.statistics,arrows,backgrounds}
\pgfplotsset{compat=1.14}
\graphicspath{{figures/}}

\usepackage[backend=biber,style=ieee,maxbibnames=6,minbibnames=6]{biblatex}
\newcommand{\etal}{\emph{et al.}\xspace}

\newtheorem{definition}{Definition}

\begin{document}

\title{Constraint-Guided Reinforcement Learning: Augmenting the Agent-Environment-Interaction}
\author{%
\IEEEauthorblockN{Helge Spieker}
\IEEEauthorblockA{\textit{Simula Research Laboratory} \\
Fornebu, Norway \\
helge@simula.no}\thanks{This work has received funding from the European Union under grant agreement no. 825619 (AI4EU).}}

\maketitle

\begin{abstract}
Reinforcement Learning (RL) agents have great successes in solving tasks with large observation and action spaces from limited feedback. 
Still, training the agents is data-intensive and there are no guarantees that the learned behavior is safe and does not violate rules of the environment, which has limitations for the practical deployment in real-world scenarios.
This paper discusses the engineering of reliable agents via the integration of deep RL with constraint-based augmentation models to guide the RL agent towards safe behavior. 
Within the constraints set, the RL agent is free to adapt and explore, such that its effectiveness to solve the given problem is not hindered. 
However, once the RL agent leaves the space defined by the constraints, the outside models can provide guidance to still work reliably.
We discuss integration points for constraint guidance within the RL process and perform experiments on two case studies: a strictly constrained card game and a grid world environment with additional combinatorial subgoals.
Our results show that constraint-guidance does both provide reliability improvements and safer behavior, as well as accelerated training.
\end{abstract}

\section{Introduction}

Autonomous systems with self-adaptive or learned components have applications in a variety of tasks due to the increasing interest in industrial automation, industry 4.0, or cognitive robotics.
However, for the trustful application of these systems reliability and safety are crucial factors.

Reinforcement Learning (RL) has been successfully applied to control a variety of applications~\cite{Arulkumaran2017}, such as game playing~\cite{Justesen2019}, robotics control~\cite{Kober2013}, or combinatorial optimization~\cite{Dai2017}. %
The RL agent explores its environment by trying random actions and learns through reward signals about its dynamics. 
Through sufficient exploration and discovering successful strategies, it converges towards an optimal policy and solves the problem that is defined by the environment and its reward function.
As an RL agent learns from trial-and-error, there is no possibility to employ the agent with hard rules or safety guidelines from the beginning on.
And also after training, there are no given guarantees for the safe and reliable behavior of the agent.
Furthermore, training RL agents is often data-inefficient and takes many iterations of training runs in sometimes computation-intensive simulations or even more costly on actual physical devices.
These properties of RL agents make it challenging to deploy them, either as stand-alone agents or as trainable components within a larger system architecture~\cite{Dulac-Arnold2019}.
Their integration requires handling uncertainties and to ensure they are applied within their designated environment in a more strict way than common software components.
The RL agent is not aware that it is receiving inputs that it is not sufficiently trained for and will be select an action in any case even without having been in the same or similar state before.
An exception to this is Bayesian RL \cite{Ghavamzadeh2015} where uncertainty estimations are considered, but these algorithms have not yet reached the same level as the more common model-free RL algorithms we consider.

Contrary to RL, symbolic AI, such as expert systems or constraint solving, can make decisions based on an explicit constraint model of the environment, but depending on its complexity, it is often infeasible to model and solve the environment for every decision. 
Still, constraint models are data-efficient, as there is no training, and always avoid unsafe behaviors as defined by the model.

In this work, we make an argument for the integration of both techniques, such that the RL agent, also the \emph{learner}, is guided by an encapsulating constraint model that describes safety constraints for the agent's task.
We refer to this integration as \emph{constraint-guided reinforcement learning} (CGRL) and it has the goal to design a reliable and safe agent.
Within this hybrid agent, safety models are used to augment the interaction of the RL agent with its environment to avoid unsafe or undesired actions and behaviors. 
More specifically, we discuss three interfaces to augment the agent-environment interaction: 
1) the interface between environment and agent, where external state information about the environment is observed; 
2) the external interface between agent and environment, where selected actions are transferred to the environment; 
3) the internal decision-making within the agent, to directly adjust which actions are taken.

Encapsulating RL agents within an outside model for safety or training purposes has been explored before \cite{Garcia2015}, e.g. in the context of shielding reinforcement learning agents~\cite{Alshiekh2018} or imitation learning, where the agent is shown the behavior of experts and tries to mimic them~\cite{Price2003,Ho2016}.
Our focus for the integration within the scope of this work lies on the different interfaces for augmentation and their effects on the system's design and reliability.
We further discuss constraint solving techniques to model acceptable behavior or to define how to act in case of bad decisions from the RL agent.

Developing dedicated encapsulation techniques, represented by the different constraint models on the interfaces, allows setting strict rules and guidelines, while not prohibiting the free exploration and decision making of the agent within the set boundaries.
To underline the relevance and motivate further research in this direction, we discuss the general working mechanisms of reinforcement learning as well as the potential integration points for an external constraint model. 
The implementation of CGRL and its effectiveness is shown in two case studies. 
Our results confirm that constraint-guidance is beneficial for both faster training and better final performance than pure model-free reinforcement learning.

The main contributions of this paper are:
1) We integrate safety models in the RL process at the intersection between agent and environment to guide the decision-making.
2) We present an architecture to implement these models by using both satisfaction and optimization techniques.
3) We perform an experimental evaluation of the presented method on two case studies: a card game and a grid world with a combinatorial puzzle goal.

The remainder of the paper is structured as follows:
After reviewing the necessary background on RL and constraint solving as well as existing works on their intersection in Sections~\ref{sec:background} and \ref{sec:relwork}, we discuss their integration for constraint-guided RL in-depth in Section \ref{sec:integration_cp_rl}.
In Section~\ref{sec:experiments}, we implement CGRL in two environments and evaluate its effects. 
Before concluding the paper in Section~\ref{sec:conclusion}, we discuss the impact and current challenges for CGRL as well as opportunities for future research in Section~\ref{sec:discussion}.
 
\section{Background}
\label{sec:background}

\subsection{Reinforcement Learning}
\label{sec:rl}

In reinforcement learning (RL) an agent interacts with an environment by observing the environment's state $s_t$ at time step $t$ and selecting an action $a_t$. 
From the effect of the action, a reward $r_{t+1}$ is received and a new state $s_{t+1}$ is reached. 
The goal of the RL agent is to maximize the expected return, described as the cumulative discounted reward at time $t$, that is, $G_t = \sum_{k=0}^{\inf} \gamma^k r_{t+k+1}$, where $\gamma \in [0, 1)$ is the discount factor.
More formally, the environment is characterized by the tuple $\langle S, A, T, R \rangle$, with $S$ being the set of states in the environment, $A$ being the set of actions.
$T : S \times A \rightarrow S$ is the transition function from the execution of an action in one state leading to another state and $R$ is the reward function, which assigns a numerical feedback to an action in a state $r : S \times A \rightarrow \mathbb{R}$.

The agent follows a trainable policy $\pi$ from which it chooses the action in each state $\pi : S \times A \rightarrow [0, 1]$.
The policy assigns each action $a$ a probability to be taken and from this probability distribution either the action with highest probability is picked greedily or it is randomly sampled to encourage exploration.
This general agent-environment interaction process is shown in Figure \ref{fig:integration}, including extensions for CGRL which we will discuss at a later point.
The agent interacts with the environment in finite episodes, which means that some states are terminal states and once these states are reached, the episode is over and a new episode starts. Terminal states can represent both desired goal states, e.g. in a maze where a certain location has to be reached, but also undesired states, e.g. the agent performs illegal or unsafe behavior.
For a detailed introduction to RL, we refer the interested reader to the book by Sutton and Barto (2018).

Over the recent years, deep RL, where the action selection policy is approximated by deep neural networks, has received large attention and many RL algorithms have been proposed in the literature \cite{Arulkumaran2017}, for example, Proximal Policy Optimization (PPO) \cite{Schulman2017} which we use throughout the experiments in this paper.
These algorithms share, that they represent the state as input to the neural network and receive a vector of action scores as output.
These scores are interpreted as probabilities and the action is randomly sampled or greedily chosen by the highest score.
The state representation is usually problem-specific. 
For problems where the state resembles a form of spatial information, e.g. images, convolutional neural networks (CNN) are commonly used, whereas otherwise fully-connected feed-forward neural networks are common.

We use PPO in an actor-critic way~\cite{Mnih2016}, where the agent is modeled by two neural networks.
The first network, the actor, outputs the scores for each action as described above.
The second network has only a single output that predicts the expected reward for the current state.
Both networks usually share the same architecture and also the weights of the initial layers with the only difference being the final layers, where the actor has as many outputs as actions, the critic only has a single real-valued output.
Using a critic network helps to stabilize and accelerate the learning process since it allows to compare the current policy against the earlier made experiences and supports the improvement of the policy.

\subsection{Constraint Solving}

Throughout this paper, we refer to constraint solving in the context of modeling the acceptable behavior of a learner, like the RL agent, which is either a satisfaction problem, when asking "Does the learner behave according to the specification?", or an optimization problem when asking "What is the state representation or action that complies with the specification with only minimal changes?".
Modeling a satisfaction problem allows to identify undesired states and actions, but does not guide the learner, except when designing each individual action as separate problems.
The optimization model instead allows augmenting the state or action towards specification compliance, according to its constraints.

Formally, we define a \emph{constraint satisfaction problem} (CSP) as a triple $\langle \mathcal{X},\,\mathcal{C},\, \mathcal{D} \rangle$ where $\mathcal{X} = \{\mathbf{x}_1, \dots, \mathbf{x}_n\}$ is a set of variables, $\mathcal{C}$ is a set of constraints $\mathcal{C}=\{c_1,\dots,c_m\}$, where each constraint $c_i$ involves a subset of the variables.
Each variable $\mathbf{x}_i$ is also associated with a finite integer domain $\mathcal{D}(\mathbf{x}_i)$ of all its possible values.
A variable assignment $\theta$ satisfies the CSP $\langle \mathcal{X},\,\mathcal{C},\, \mathcal{D} \rangle$, if each variable $x$ has a value assigned from its domain $D(\mathbf{x}_i)$, such that none of the constraints $\mathcal{C}$ are violated.
A \emph{constraint optimization problem} (COP) extends the CSP formulation by an objective function $f_z$, which is calculated over a subset of the variables, with an objective value $z$, which, in our context, is to be minimized, while the variable assignment $\theta$ still satisfies the constraints $\mathcal{C}$.
For a detailed overview on constraint solving, we refer the interested reader to \cite{Rossi2006}.

The constraint model can be modelled and solved by dedicated constraint solving libraries, e.g. MiniZinc or Essence. %
This has the advantage of having solver-provided guarantees for the correctness of results - under the assumption of a correct model - and tools optimized for the task, but with the disadvantage of increasing system complexity due to additional libraries and related dependencies.
The alternative would therefore be to implement the constraint model within the programming paradigm of the autonomous system.
In this case, there are no additional external dependencies, but the implementation is potentially more error-prone since constraint modeling languages are usually more expressive for the tasks than general-purpose programming languages.
We will come back to the discussion between these two alternatives in the discussion part of the paper in Section~\ref{sec:discussion}.

\section{Related Work}
\label{sec:relwork}

The main areas of related work are safe, ethical, or social reinforcement learning and the design of autonomous systems.
The work closest to ours is the method of \emph{Shielded Reinforcement Learning} by Alshiekh~\etal~\cite{Alshiekh2018} .
The authors propose a safety framework where a shielding component is deployed either between the environment and the agent or after the agent has made its action selection. 
The shield represents a safety specification, based on temporal logic, that assures the correctness of the system.
This is related to our method, which also considers integrating a constraint-guidance component at the same interfaces.

Similar works have been based on using barrier functions~\cite{Cheng2019} or provable formal theories of the acceptable behavior specifications of the RL agent~\cite{Fulton2018}, or focus on specifics of the RL environment, e.g. stochastic dynamics in robotics~\cite{Li2020a}. 
These approaches focus mostly on the provable aspects of the formalization, and strongly motivate the necessity and feasibility of hybrid approaches, but focus less on the aspects of how they are implemented and expressed.
A further avenue is the design of RL methods that have safety inherently in their functional methodology, e.g. \cite{Achiam2017,Chow2018}. 
Unlike these methods, we provide safety extensions to any form of RL agent, even without inherent safety notion. 
We see an advantage in a modular approach, however, we do not argue that one or the other is necessarily better.
Finally, for an in-depth overview on safe reinforcement learning, we refer to the survey by García and Fernández~\cite{Garcia2015}.

Understanding and steering how agents behave and learn in unknown environments also affects ethical concerns and is an aspect of active research to include ethical standards as boundaries for the agents' behavior~\cite{Rossi2019}.
For example, Noothigattu~\etal explore the effects of imposing ethical constraints on a Pacman agent, such that it is prohibited to eat ghosts~\cite{Noothigattu2018}.
The consideration of ethics for learning agents is of relevance and imposing constraint-guidance, based on models that capture ethical constraints, proposes an approach towards this challenge.
Another area of interest is to let the agent learn which actions are either not safe, not desired, or just unfeasible in a given state. 
Zahavy~\etal~\cite{Zahavy2018} propose a method where the agent learns which actions not to choose in a given state. 
Learning when an action is infeasible and should be avoided or when it is just not available has been part of teaching an agent's policy, where the agent learns which actions to avoid, but without an explicit specification of wrong and forbidden actions.

\section{Integrating Constraints and RL}
\label{sec:integration_cp_rl}

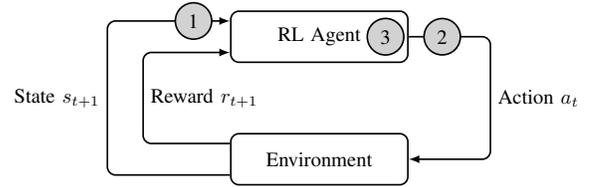
\begin{figure}[t]
    \centering
    \resizebox{0.9\columnwidth}{!}{\tikzstyle{block} = [rectangle, draw, 
    text width=8em, text centered, rounded corners, minimum height=2.5em]
    
\tikzstyle{line} = [draw, -latex, rounded corners]

\begin{tikzpicture}[node distance = 6em, auto, thick]
    \node [block] (Agent) {RL Agent};
    \node [block, below of=Agent] (Environment) {Environment};

    \path [line] (Agent.0) --++ (4em,0em) |- node [near start]{Action $a_t$} (Environment.0);
    \path [line] (Environment.190) --++ (-6em,0em) |- node [near start] {State  $s_{t+1}$} (Agent.170);
    \path [line] (Environment.170) --++ (-4.25em,0em) |- node [near start, right] {Reward $r_{t+1}$} (Agent.190);
     
    \node [draw, circle, fill=gray!35, text centered, left=0cm and 0.3cm of Agent.170] (ObservationMask) {1};
    \node [draw, circle, fill=gray!35, text centered, right=0cm and 0.25cm of Agent] (ActionReplacement) {2};
    \node [draw, circle, fill=gray!35, text centered, left=0cm and 0.07cm of Agent.0] (AgentInternal) {3};
\end{tikzpicture}}
    \caption{The Agent-Environment Process with constraint-guidance integration: (1) Environment-Agent, (2) External Agent-Environment, (3) Internal Agent-Environment.}
    \label{fig:integration}
\end{figure}

\begin{figure*}[t]
    \centering
    \subfloat[Environment-Agent]{%
        \includegraphics[width=0.38\textwidth]{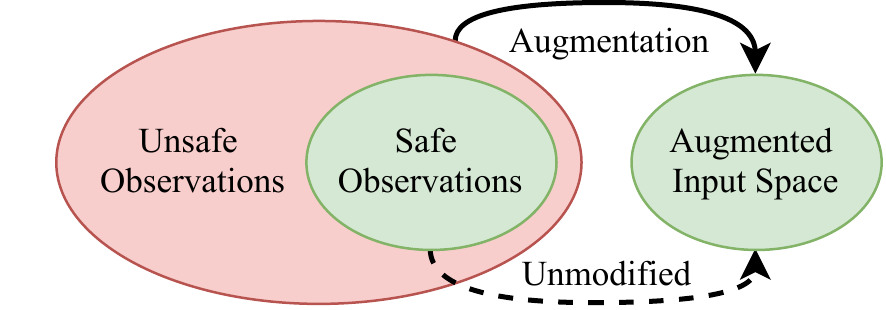}}\hfill
    \subfloat[External Agent-Environment]{%
        \includegraphics[width=0.3\textwidth]{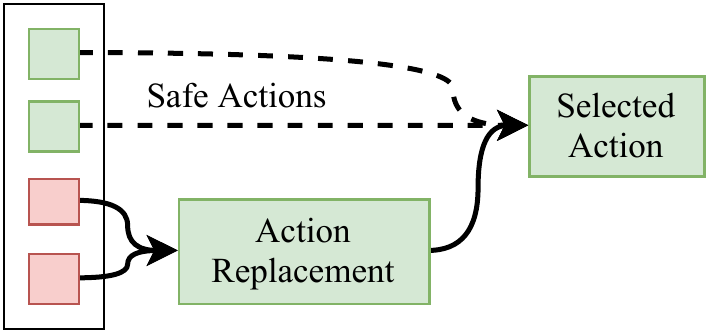}}\hfill
    \subfloat[Internal Agent-Environment]{%
        \includegraphics[width=0.3\textwidth]{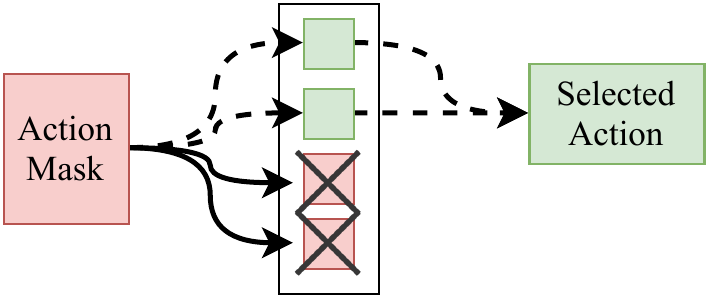}}
    \caption{Effects of the augmentation interfaces on the observation and action space. The Environment-Agent interface augments the observation space to map unseen or unsafe observations to a familiar environment in which the agent can safely act. The Agent-Environment interfaces replace unsafe actions by a safe default (External) or limit the availability of actions for the agent before selection (Internal). The safety of an action depends on the observation and can vary from step to step.}
    \label{fig:space_reduction}
\end{figure*}

To achieve constraint-guided reinforcement learning, the constraint model needs to be integrated within the agent-environment process.
In this section, we identify and discuss the three interfaces for constraint-guidance (Figure~\ref{fig:integration}) and their impacts on the functionality and the autonomy of the encapsulated RL agent (Figure~\ref{fig:space_reduction}).

One integration point lies at the interface between the environment and the agent, where observations and rewards are received. 
The second integration point addresses the interaction from the agent to the environment, i.e., when an action is issued by the agent. 
These two interfaces wrap the RL agent as a black-box and manipulate the inputs and outputs of the agent. 
The third integration point, on the other hand, lies within the agent and allows to apply constraint-guidance in the internal mechanisms of the RL agent.
Relating to the existing literature, the first two interfaces are similar to the work by Alshiekh \etal \cite{Alshiekh2018} when their proposed shielding is introduced either before or after the action selection part.
The constraint models for each of the interfaces are modeled based on a risk- and safety-analysis of the system.
The approach is to formalize the most essential boundaries for the agent's behavior without restricting the agent's freedom within these boundaries more than necessary.
We discuss each of the three integration types in the following and consider the interfaces that have to be implemented by a constraint-guidance mechanism.

\subsection{Environment-Agent Interface}

The \textit{Environment-Agent Interface} augments the transfer of state information from the environment to the agent in a form of state preprocessing.
By augmenting the state information, either additional information can be added to the state, or parts of the state can be masked or modified to avoid unsafe decisions.
One application of this is to reduce the decision space for the RL agent.
Aspects of the state, which would lead the agent to act in a specific, but potentially dangerous way, are replaced by safe dummy values or a representation that does not lead to a certain action, but also does not misrepresent the actual state space.
For example, when having an agent with an inventory of items, those items which are dangerous to be used in the current state can be omitted from the augmented state that is given to the agent.

\begin{definition}[Environment-Agent Interface]
The interface function $f : S \rightarrow S$ augments the observed state, such that the difference diff($s_t, s'_t$) is minimized while following side constraints $\mathcal{C}_{Env}$, and returns a modified state $s'_t$.
\end{definition}

Technically, the augmentation component is based on a constraint model $c(s_t) = s'_t$, that specifies potentially dangerous states, and can make modifications
$c(s)$ receives a state $s_t$ and returns a state $s'_t$, which can be identical to the original state if $s_t$ is not considered dangerous, or modified to exclude or mask the undesired parts of the state.

To implement this interface in an \emph{Observation Mask} model, that augments the observed state for the agent, a constraint optimization model can be used.
The instance parameters are the observed state, the decision variables are the augmented state, and the objective function measures the changes between the observed and augmented state.
During optimization, the augmented state is adjusted such that the state complies with the modeled specification, e.g. unusable items from an agent's inventory are masked while making as few modifications as possible.
Making only minimal changes is relevant to keep the augmented state close to the actual observation because large changes can lead the agent to make an unsuitable decision for the actual state and thereby hinder the success within the environment.

\subsection{External Agent-Environment Interface}

Once the agent selected an action, this action is passed back to the environment to be executed.
At this point, the \emph{Agent-Environment Interface} augments the selection and can interfere by observing both the most recent state and the selected action in order to decide whether the chosen action is acceptable or violates some guidance criterion, e.g. being unsafe or potentially harmful.
In case the action is not accepted, the interface can replace the selected action following a heuristic procedure.
This replacement action might not be optimal as it does not follow the learned strategy but can avoid unsafe outcomes instead.

\begin{definition}[External Agent-Environment Interface]
The interface function $f : S \times A \rightarrow A$ receives the last observed state $s_t$ and the taken action $a_t$ as input and returns a, probably modified, action $a'_t$. 
Action $a'_t$ must obey the constraints $\mathcal{C}_{Env}$ for the environment and the last observed state.
\end{definition}

An \emph{Action Replacement} model can be implemented either as a satisfaction model, that validates whether an action follows the specification and replaces invalid actions by a default action or as an optimization model that has a notion of distance between actions and can optimize the action change to be minimal while following the specification.
The Agent-Environment Interface forms thereby a fallback strategy that monitors the RL agent's behavior and only if it does not behave according to the specified acceptable behavior, its decisions are overruled.

\subsection{Internal Agent-Environment Interface}

The first two augmentation interfaces are decoupled from the RL agent and see it as a black-box.
The third interface, \emph{Internal Agent-Environment Interface}, is integrated within the action selection mechanism of the RL agent.

While augmenting the observation and checking the selected action against a constraint model allow to prohibit undesired or unsafe behavior, the RL agent is not aware of this constraint-guidance and does not learn that its action was overruled by an external mechanism. 
That means the agent does not learn to avoid this action the next time the same state is encountered and the observed reward does not match the selected action.
The internal integration into action selection overcomes this, but at the cost of requiring access inside the agent, unlike the previous interfaces.

\begin{definition}[Internal Agent-Environment Interface]
The interface function $f : S \rightarrow \mathbb{R}^{n}$ where $n$ is the number of available actions: $f(s_t) = m$.
Each action is assigned a masking score, which is inserted into the policy $\pi$ and multiplied with the network outputs before selecting an action.
From the masked output, the action selection is continued with a bias towards actions favored by $f$.
\end{definition}

Using constraint-guidance, the scores of the individual actions can be adjusted to either discourage unwanted actions by decreasing their score or to encourage the selection of desired actions by increasing their score.
In the simplest case, $m$ is a boolean vector where each action is masked as allowed/available or unavailable for the current state.
Any of these adjustments steer the RL agent towards selecting desired actions for which then a reward is received and the policy is trained with the normal training procedure.

An \emph{Action Mask} model, that implements the internal agent-environment interface, can be interpreted as a constraint satisfaction model with one decision variable per action, where a solution enables all allowed actions for the given state.

\subsection{Deriving Constraint Models}
\label{sec:deriving_constraints}

All interfaces require the existence of a constraint model for augmentation.
There are different ways to derive these constraint models and how to formulate them.
The first option, which we consider mostly in this paper and for the experimental evaluation, is to have the constraint model handcrafted from expert knowledge.
This knowledge can either stem from precisely analyzing error causes and avoiding these situations or setup generic rules for unsafe states and actions that are known to cause unrecoverable error situations.

The second option targets to iteratively refine the constraint model through a dedicated learning process, for example using constraint acquisition or learning~\cite{Bessiere2015}.
This process would require access to a safe simulation environment as the initial set of rules cannot be expected to be safe.
Over subsequent iterations, the RL agent can act using the initial constraint model, and, in case of unsafe situations and error cases, the provoking states and actions are used to refine the constraint model via a constraint learning step.
We do not further consider this option in this work, but leave it as a promising direction for future work on CGRL.

\section{Experiments}
\label{sec:experiments}
In the following section, we apply constraint-guided reinforcement learning in two environments.
The structure for each case study environment follows the same approach.
At first, we introduce the problem setting, the available state information for the RL agent, and the reward structure.
Afterwards, we discuss the constraint-guidance models to integrate with the RL agent.
In each environment, we consider three constraint-guidance models, one for each of the integrations previously discussed, and a vanilla RL agent without constraint-guidance as a baseline for comparison.

\begin{figure}[t]
    \centering
    \includegraphics[width=0.9\columnwidth]{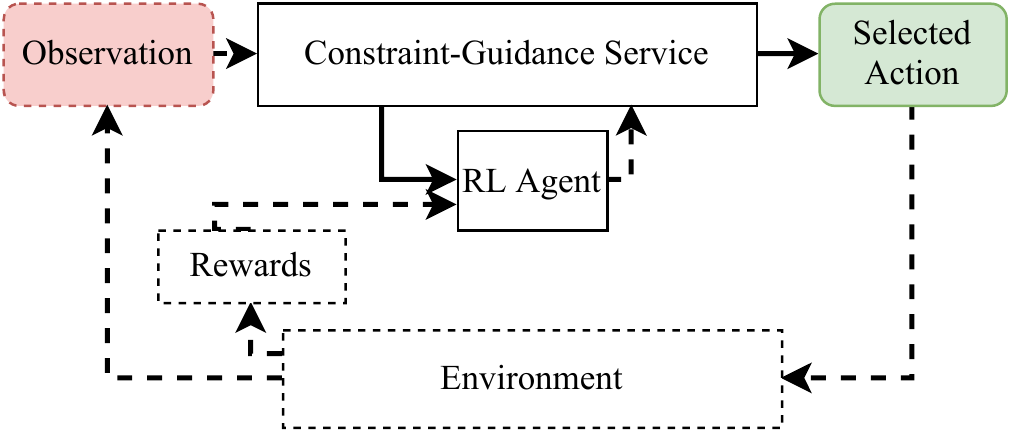}
    \caption{Architecture: The constraint-guidance model acts as a wrapper and has first access to observations and final control over the selected actions.}
    \label{fig:architecture}
\end{figure}

The RL algorithm is PPO (Proximal Policy Optimization) \cite{Schulman2017}\footnote{The PPO implementation is available here: \url{github.com/lcswillems/torch-ac}}. 
All runs use the same hyperparameters, which closely follow the defaults proposed by the authors, except for the network architecture, which is adjusted for each environment.

\subsection{Environment I: Card Game}
\label{sec:thegame}

\subsubsection{Scenario \& Rules}
In the first experiment, we focus on a card game, modeled after ``The
Game''\footnote{\textit{The Game: Are you ready to play The Game?} (2015). Created by Steffen Benndorf. Published by N{\"u}rnberger-Spielkarten-Verlag.}.
The goal of the game is to distribute 98 unique cards, ranging from 2 -- 99, onto four card piles.
Two of these card piles are to be filled in increasing order from 1 -- 100, the other two in decreasing order from 100 -- 1. 
It is only allowed to place cards with a higher respectively lower card value on each of the piles, otherwise the action is invalid.
There is one exception to this rule. 
A lower respectively higher card value can be played if the difference in card value is exactly 10, i.e. if the increasing pile has card 42 on top, card 32 is allowed on top of it.

The game ends when all cards have been distributed or there is no further move possible.
The original card game can be played with 1--5 players, in this case study we focus on the single player version, where the player has 8 cards on their hand and refills their hand from the stack after having played two cards.

\subsubsection{Environment \& Agent}

The state is represented by a 12-di\-men\-sio\-nal vector, consisting of the four card piles and the eight hand cards, sorted in increasing order.
If a slot for a hand card is empty, because a card was played and no new cards were drawn yet, it is replaced by a placeholder value of $-1$.
The agent can place any hand card on any of the card piles, leading to a total of $8 \times 4 = 32$ possible actions.
The agent's policy is represented by a feed-forward neural network with three fully connected layers of each 32 hidden nodes with tanh activations.
Both the actor and the critic network share the weights of the initial layers.

The agent receives a positive reward of $r = 0.1$ for each valid move it plays and a negative reward of $r = -0.1$ for each invalid move.
At the end of the round the final reward equals the number of played cards, i.e. at most 98.

\subsubsection{Constraint-Guidance Models}

We define the CGRL models such that invalid actions are avoided, but do not impose are restrictions on the inner RL agent.
If an action has to be selected through constraint-guidance, it greedily picks that action which has the smallest difference between the card piles and the hand card value.

\textbf{Observation Mask}
The Environment-Agent Interface is implemented such that cards with only negative effects, i.e. they cannot be played on any of the card piles, are hidden from the RL agent by showing that these card slots are empty.
If all cards can be placed on at least one of the piles, no changes are made.

\textbf{Action Replacement}
The External Agent-Environment Interface monitors the selected action for invalid card placements, which would result in a negative reward penalty. These actions are heuristically replaced by that action which incurs the smallest step on any of the card piles, i.e. that card which the minimum difference between card and pile value.

\textbf{Action Mask}
The internal environment-agent interface behaves similar to action replacement, but instead of replacing the action after it has been selected, it masks those actions which are invalid.
This enforces the agent to always select a valid action and focuses the exploration towards understanding good game strategies without having to learn about the validity of actions.
Further variants can be considered, for example, it is possible to enforce playing cards where the special rule of decreasing/increasing the card pile in the "wrong" direction can be used. 
But, this goes beyond the usage of constraint-guidance for safety aspects and enforces a particular playing strategy upon the agent, which would be less comparable to the other models.

\subsubsection{Results}

\begin{figure}[t]
    \centering
    \resizebox{0.8\columnwidth}{!}{\input{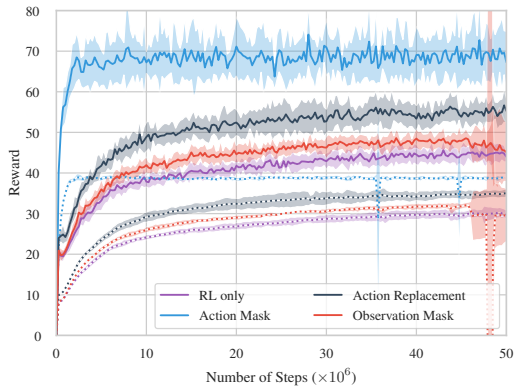}}
    \caption{Card Game: Maximum (solid lines) and average (dotted) rewards with and without masking. We show the mean and std. dev. over 10 training runs.}
    \label{fig:thegame_training}
\end{figure}

The reward curves of the four different agents over a training time of 50 million frames are shown in Figure~\ref{fig:thegame_training}.
We train the RL agent such that it runs multiple environments in parallel to collect experiences for training.
This rollout both allows to run the training in parallel on multiple CPU cores as well as it stabilizes it, because more diverse states and rewards are available.
We show both the best and the average rewards of 64 training environments.

The results show a strong impact of constraint-guidance over the RL-only agent, but they also show differences between the different integration techniques.
The best performance is reached with Action Mask model, which implements the internal agent-environment interface, and manipulates the available actions by removing invalid actions.
Its performance surpasses that of the other agents from the beginning and quickly reaches a high level.

The other models follow a similar learning curve but are shifted in their base performance, which we attribute to the effects of the different augmentation models.
Augmenting the observations, i.e. removing hand cards that are not possible to be played in the current state, is the least beneficial mask in this scenario, and also causes instability in the training process visible by the high fluctuation at around 49 million frames.
This is caused by the game mechanics, where only in few cases a card is actually invalid to be played on all four card piles. It is much more common that a card is only invalid for a few piles, but this cannot be modeled in our setup by augmenting only the observation, but requires adjusting the actions, like the Action Mask does it.
Still, it can be helpful and performs better than RL only.

Regarding the training duration until the agent's performance converges and does no longer improve by a large margin, we observe faster convergence when using the action mask model compared to the external augmentation.
The RL only and the Observation Mask models steadily improve over a longer time, but their main performance level is reached in a similar time than the others.

\subsection{Environment II: Grid World Puzzle}
\subsubsection{Scenario \& Rules}

\begin{figure}[t]
    \centering
    \includegraphics[width=0.22\textwidth]{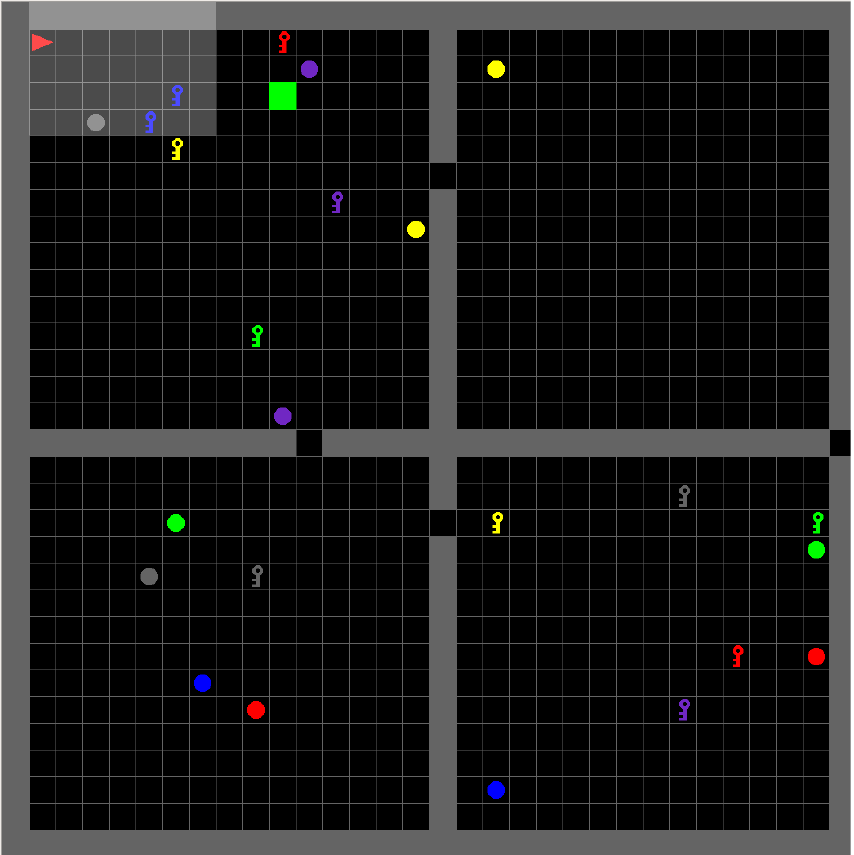}
    \caption{Grid World Puzzle: The agent (red arrow with observable space highlighted) has to collect 16 items (circles and keys in 8 colors) before reaching the goal (green square).}
    \label{fig:grid world_environment}
\end{figure}

The second environment is a puzzle variant of the grid world problem.
The agent is placed in a random 2D grid world environment, consisting of four rooms, and has the goal to collect a number of distinct items before finding the exit door of the grid world (see Figure~\ref{fig:grid world_environment}).
It can move forward, turn left or right, or pickup an item that is in front of the agent.
Every item is described by its shape (key or circle) and its color.
The goal is to pick one item of each type and reach the exit door in as few steps as possible.

This environment poses two challenges for the agent:
1) The visible state of the grid world is limited to the eight surrounding tiles next to the agent's position, which requires exploration of the environment; 
2) the agent needs to learn to avoid items that it has already collected. 
The inventory of already collected items is part of the agent's state and given to the agent at each step.

\subsubsection{Environment \& Agent}

The total grid world environment is $32 \times 32$ tiles, divided into four equally sized rooms connected with open doors. 
Both the placement of doors and the items is randomly generated for each new environment.
The state of the environment is represented by the agent's field of view, $7 \times 7$ fields, and its current inventory, listing all collected items.
Grid world tiles can be either a plain field, a field with an item, or the exit door, all of which can be accessed by the agent at any time, or a wall, which blocks the agent and cannot be accessed.
The set of actions is $a \in \{\text{forward}, \text{left}, \text{right}, \text{pickup}\}$.
The agent's policy is represented by a small convolutional neural network with three 2D convolutional layers followed by two fully connected layers.
Each layer is connected with ReLU activation functions.
Again, both the actor and the critic network share the weights of the initial layers.

The reward function is modeled such that the agent receives a positive reward of $1/16$ for each distinct object it collects.
Collecting a duplicate item, that is already in the inventory, causes a negative reward of $-1$ and ends the episode.
The maximum length of an episode are $8192$ steps, once this number of steps is reached, the episode is also terminated.
When the exit door is reached, the agent receives an additional reward of $1 - 0.8 \times \text{Number of steps}/8192$.
This reward function does only return sparse rewards at specific events instead of continuously giving feedback to the agent, which is contrary to the card game environment, where every action of the agent is rewarded by the environment.
The implementation of this environment is based on the \emph{Mini\-Grid} project \cite{gym_minigrid} with extensions for the additional puzzle objective.

\subsubsection{Constraint-Guidance Models}

The goal of the constraint-guidance models for the grid world puzzle environment is to avoid collecting duplicate items, i.e. the undesired behavior.%

\textbf{Observation Mask}
The model for augmenting the observed states from the environment hides any duplicate item from the observed state and masks it as a wall.
This signals to the agent, that the masked tile is not of interest and encourages the agent to further explore or focus on another, unmasked item.

\textbf{Action Replacement}
When the agent decides to move to a tile with a duplicate item, this model observes this specification violation and replaces the move by turning right.

\textbf{Action Mask}
To avoid picking duplicate items, the action mask disables the action to pick an item as long there is no distinct new item in front of the agent.
Thereby the agent can either move, as long as there is no new item in front of it, or it is set to pick a new item, which disables the movement.
This guidance model removes the logic for picking items from the learned agent, which is then only concerned with navigation towards distinct items.

\subsubsection{Results}

\begin{figure}[t]
    \centering
    \resizebox{0.8\columnwidth}{!}{\input{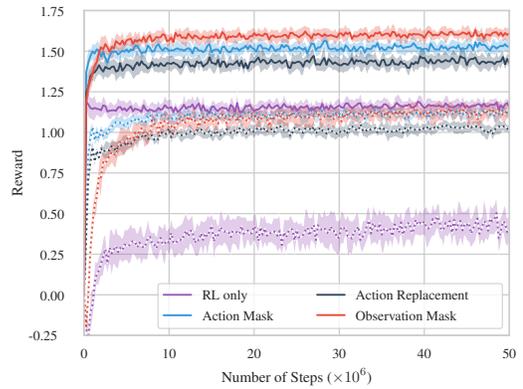}}
    \caption{Grid World Puzzle: Maximum (solid lines) and average (dotted) rewards with and without masking. Shown is the mean and standard deviation over 10 training runs.}
    \label{fig:gridworld_training}
\end{figure}

The results, in form of training curves, are shown in Figure~\ref{fig:gridworld_training}.
As in the previous experiment, we trained the agents with 64 environments and parallel and report both the best and average performance over time, repeated with 10 different random seeds.

Models using constraint-guidance perform better than the RL-only agent, with the average CGRL performance being competitive to the best rewards of the RL-only.
This results show the general difficulty of the environment for a standard RL algorithm.
The average reward is close to the minimum reward that is received for successfully reaching the exit door, but does not show signs for increased attention towards successfully collecting unique items. 
The CGRL models on the other hand achieve higher rewards, with the Observation Mask model, that excludes already collected items from the observed state, performing best before the Action Mask and Action Replace models.
We further observe a faster performance increase in the Action Mask model than in the eventually better Observation Mask model.
While the observation mask excludes already collected items from the observed state by presenting them as walls, the agent still can walk towards them and collecting them, as the selected action is not augmented. As the agent needs to learn that walls cannot be accessed, the observation mask is not as effective as the action mask in the beginning.
The action mask, which was most effective in the card game, disallows the agent to access fields with already collected items, but only when the agent stands directly in front of the item. Before that state is reached, the agent can still plan to collect the item, if it has not learned to avoid duplicates. This can cause the lower performance of this mask.

\section{Discussion}
\label{sec:discussion}

Both performed experiments showed the effectiveness of CGRL for guiding a RL agent in complex environments.
This result confirms our expectations, since we introduce external domain knowledge into the training process and the agent's behavior.
However, we also observe a better data-efficiency of the training and the avoidance of unsafe behavior, as confirmed by higher rewards.
Better data-efficiency is important when a single step of the RL agent is more costly than in the small and computationally cheap environments used here, e.g. when running a complex physical simulation or even using physical components, there it is especially useful to model the acceptable behavior via a constraint model.
In addition to the physical environment, where unsafe behavior might be harmful for the machine or the physical environment, the avoidance of unsafe behavior is crucial for the application and deployment to either safety-critical or ethically challenging environments.

A challenge in the application of CGRL can occur from the additional runtime overhead added from the constraint solving for augmentation.
As the model has to be run for every step of the RL agent a small overhead can accumulate to a longer runtime.
However, this limitation mostly applies only in cases where the overhead is significant in relation to the general computational cost of one step in the environment. 
In an environment where every step is already costly, the benefits of faster training and less undesired behavior will exceed the disadvantage of a runtime overhead.

As future work on CGRL, besides the challenges raised, we plan to extend the method to environments with more complex state representations.
Here it can be necessary to additionally preprocess the observation before constraint reasoning can be applied.
Another direction is to adopt principles from imitation learning, where expert behavior can be observed, to learn augmentation rules, e.g. via constraint acquisition \cite{Bessiere2015}, as addressed in Section~\ref{sec:deriving_constraints}.
This way, CGRL would first learn a static constraint model for guidance and then allow an RL agent to learn a strong policy within its specification.

\section{Conclusion}
\label{sec:conclusion}

We presented Constraint-Guided Reinforcement Learning (CGRL), a method to combine constraint solving and reinforcement learning into a hybrid agent that behaves according to a specification, but is otherwise free to explore and learn the best policy for the problem.
CGRL augments the interfaces between the RL agent and the environment through constraint models, which specify acceptable behavior that the agent should comply to. 
Constraint solving acts here as a safety fallback for the encapsulated RL agent.

Augmentation can happen by changing the state observation from the environment to exclude certain aspects in order to encourage the agent towards accepted behavior, but it can also observer the actions made by the agent and adjust to avoid behavior that violates the specification. 
Finally, a third potential interface for constraint-guidance is to augment the action selection process within the agent -- unlike the first two approaches, which consider the RL agent as black-box -- and mask actions that would violate the specification.

\printbibliography
\end{document}